\documentclass[a4paper, journal]{IEEEtran}

\usepackage{subfigure}
\usepackage{color}
\usepackage{multicol}
\usepackage{multirow}
\usepackage{epsfig}
\usepackage{graphicx}
\usepackage{clrscode}
\usepackage{cite}
\usepackage{array}

\usepackage{pstricks}

\usepackage{xspace}

\usepackage{subfig}

\makeglossary

\begin{document}

\title{Efficient Non-linear Calculators}


\author{{Adedamola Wuraola and Nitish Patel}
\thanks{Adedamola Wuraola is with Scion (Rotorua) and Nitish Patel is with the Univ of Auckland.}}

\maketitle

\begin{abstract}
	A novel algorithm for producing smooth nonlinearities on digital hardware is presented. The non-linearities are inherently quadratic and have both symmetrical and asymmetrical variants. The integer (and fixed point) implementation is highly amenable for use with digital gates on an ASIC or FPGA. The implementations are multiplier-less. Scaling of the non-linear output, as required in an LSTM cell, is integrated into the implementation. This too does not require a multiplier. 
	
	The non-linearities are useful as activation functions in a variety of ANN architectures. The floating point mappings have been compared with other non-linearities and have been benchmarked. Results show that these functions should be considered in the ANN design phase.

    The hardware resource usage of the implementations have been thoroughly investigated. Our results  make a strong case for implementations in edge applications. This document summarizes the findings and serves to give a quick overview of the outcomes of our research\footnote{The authors peer-reviewed manuscripts (available at https://doi.org/10.1016/j.neucom.2021.02.030)  offer more detail and may be better suited for a thorough consideration}.
\end{abstract}


\IEEEpeerreviewmaketitle

\section{Motivation}
The General Matrix Multiplication (GEMM) typically computes $ \bar{n} = \bar{w} \times \bar{p} + \bar{b}$. In deep learning, $w$ are weights, $p$ is an input, $b$ is a bias term and $n$ is the netsum. The GEMM is the most computationally heavy workload in deep learning. The computation of the GEMM has evolved to reduce the computational complexity in terms of both memory and latency. Several optimizations and parallelization techniques have been proposed in the literature. Without a fast and resource efficient GEMM facility, Artificial Neural Network (ANN) algorithms would be dominated by the sheer volume of multiplication (MUL) instructions. With the availability of fast and resource efficient GEMM operations, the computationally complex activation functions (functions with exponent, logarithm, trigonometric, and floating-point division) become more significant and can dominate the computational load \cite{wuraola2021efficient}. Therefore, a fast and resource efficient activation function is essential to maintain the throughput of the GEMM.  

\section{\label{s:algorithm}The Algorithm}
In principle, the algorithm for generating a non-linear mapping, $f(n)$, can be written as ${f(n) = \mbox{mean}((n + r) - r)}$ where $n$ is a netsum i.e. the output of a GEMM and $r$ is a random number. The algorithm hinges on saturation of the addition operation, the saturation levels and the range of the random values. Hence, an appropriate  representation of the algorithm is given by -
 
\begin{equation}\label{eq:concept}
f(n) = \frac{1}{N} \sum _{k=1} ^N f_{sat} \left ( f_{sat} ( n + U(k), C) - U(k), M \right)
\end{equation}
Here,
\begin{itemize}
\item $n$ is a signed integer encoded with 2's complement 
\item $R$ is the bit width of the binary word used to represent $n$. The 2's complement representation implies that ${2^{R-1} \le n < 2^{R-1}}$. The output, $f(n)$ is also 2's complement encoded with a width of $R$ bits.
\item The addition and subtraction operations are saturating, i.e., 

\begin{equation}\label{eq:sat}
f_{sat}(x, Y) = \left\{
  \begin{array}{ll}
    -Y    & : x \le -Y \\
    x    & : -Y <  x < Y \\
    Y    & : x \ge Y \\
  \end{array}
\right.
\end{equation}

\item $M = 2^{R-1}$ and $0 \le C < 2^{R-2}$.
\item $U(k)$ is a predefined non-repeating sequence of integers, 
\item $N$ is the length of $U(k)$. 
\end{itemize}

The current implementation of this algorithm produces an integer non-linear mapping from an integer input.

\section{\label{s:concepts}Concepts}

\begin{figure}
\centering
\includegraphics[scale=0.9]{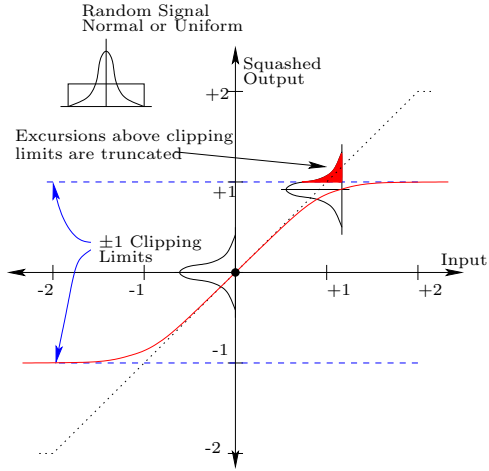}
\caption{\label{fig:description}Pictorial Description of Transfer Characteristics of the Proposed Method}
\end{figure}

\begin{figure}
\begin{center}
\includegraphics[]{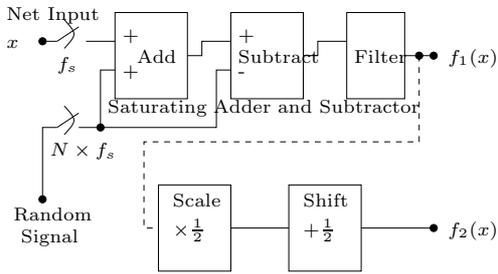}
\end{center}
\caption{Block Schematic of Proposed Square Law Squashing Function}
\label{fig:schematic}
\end{figure}

The TanSig/LogSig mappings will used to explain the concept. The technique is visualized in Fig.~\ref{fig:description} while a block diagramatic view is shown in Fig.~\ref{fig:schematic}.

First, the input i.e. the netsum and a random signal (uniform distribution with a range of $\pm1.0$) are added. The random signal is oversampled and hence the input is added to several random values. The adder is saturating (hard-limited) and hence all sums more than the defined limits are clipped. The same random signal is then subtracted from the sum. The subtraction attempts to recover the original input. If the input was small in magnitude then the addition with the random signal will not eventuate in clipping (in any of the oversampled sums) and hence the original input is recovered after the subtraction. Fig.~\ref{fig:description} shows this scenario around $(0,0)$. However, if the input is large in magnitude then, for a small subset, the addition of the random values will lead to clipping and then the following subtraction will not restore the original signal. The filter smooths the restored outputs but since a subset has reduced magnitude the filtered outputs will also be reduced in magnitude. The resulting transfer characteristic is non-linear. Fig.~\ref{fig:description} shows this around $\approx(1,1)$. If the distribution of the random signal has a very small variance then the non-linearity will be narrower i.e. it will not manifest till the signal is very close to $(1,1)$. However, if the variance is large then the non-linearity is manifest much before $(1,1)$ i.e. closer to $(0,0)$. Hence the variance can be used to shape the non-linearity. 

\begin{figure}
\centering
\includegraphics[scale=0.9]{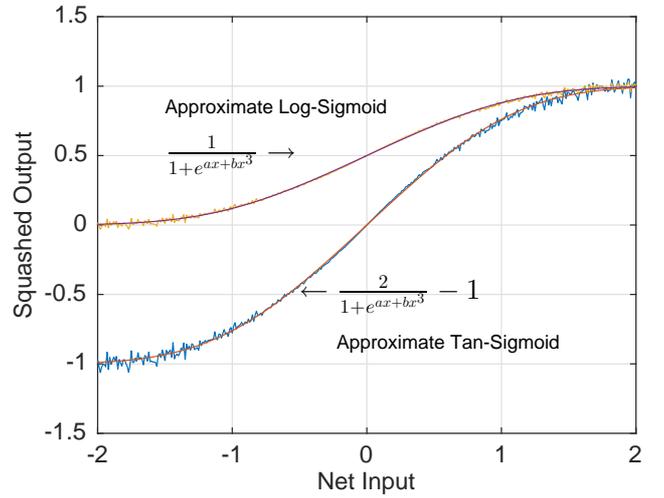}
\caption{\label{fig:sim}Simulated Squashing Functions Showing Approximate TanSig and LogSig Behavior}
\end{figure}

Fig.~\ref{fig:schematic} has been simulated in Matlab. A uniform distribution with a range of $\pm1.0$ was used. The adder has its hard-limit set at $\pm1.0$ while the subtracter is at $\pm2.0$.  The data at $f_1(x)$ was fitted to an approximate TanSig function defined in Eq.~\ref{eq:cfit_fp}. A particular Matlab simulation results in $a= -1.81$ and $b = 0.18$ (c.f. $a = -2.0$ and $b=0$ for TanSig) and exhibiting RMS error of 0.023. Thus $f_1(x) \approx f_A(x)$.
\begin{equation}\label{eq:cfit_fp}
f_A(x) = \frac{2}{1+e^{a x + b x^3}} - 1 
\end{equation}
Fig.~\ref{fig:sim} plots the Matlab simulation. The traces with the jitter are the outputs of the simulated squashing function while the solid trace is a plot of Eq.~\ref{eq:cfit_fp} with the fitted parameters.

If $f_1(x)$ is scaled and shifted by +0.5, an approximate LogSig is produced. This has also been simulated and presented in Fig.~\ref{fig:sim}.

In reality, a random source is not required required and can, in fact, be implemented using only a binary counter. Thus, in eq~\ref{eq:concept}, $U(k)$ is produced by a binary counter. Both symmetrical and asymmetrical mappings can be produced.
\section{Related work}
In literature, various approaches have been proposed for implementing efficient hardware transcendental activation functions.These methods fall into 3 broad categories: LUT, piecewise approximation, and hybrid methods. LUT is used to store the values of the activation functions as used in neurons. LUTs require pre-calculation of the activation functions mapping before being loaded into memory. As described in \cite{yang2018design}, the activation function values are divided into equal subranges, each subrange is approximated by a value stored in the FPGA read only memory. LUT method has been shown to be generally resource and memory intensive for higher resolution and precision \cite{tan2007precision}. The structure of a LUT is not amenable to direct expansion when there is a need for factors such as a change in resolution, activation function, and so on. 

Piecewise approximation of transcendental activation functions are divided into different types which include piecewise linear \cite{amin1997piecewise,larkin2006efficient}, piecewise nonlinear \cite{pushpa2014implementation}, and others \cite{tiwari2015hardware,yang2018design}. Piecewise linear approximation uses the basic form of $f(x) = ax+b$ (where a is the function slope, b is the intercept) to construct exponential-based activation functions. The authors in \cite{amin1997piecewise,larkin2006efficient} use 12 breakpoints with their solution requiring multiple IF ELSE statements, LUT to store coefficients, multiplier and adder.  The use of approximation formula as described in \cite{gomar2016precise} is based on using formula to approximate the exponential function present in exponential-based activation function as shown $e^x \approx Ex(x) \approx 2^{1.44x}$. Based on this formulation, sigmoid activation function can be calculated as $f(x) \approx \frac{1}{1+2^{-1.44x}}$. Computationally, this method requires four clock cycles to approximate the sigmoid function. The computation of sigmoid function will require an add, a subtract, a division, a multiplier and an exponent. The disadvantage of this method is high computational resource utilisation and high latency. Taylor series expansion is the most common example of piecewise nonlinear approximation for exponential-based activation functions. An advantage of Taylor series expansion is that it can be used to approximate an activation function to any precision. However, this comes at a cost of resource utilisation (several multipliers) and high latency (several clock cycles). The Taylor series expansion is of the form $f(x) = \sum _{\infty} ^{n=0} \frac{f^{(n)} (x_0)}{n!}(x-x_0)^n$. Taylor series approximation method can be combined with LUT based method and this is normally referred to as optimised/hybrid LUT. As described in \cite{yang2018design}, LUT is combined with piecewise nonlinear Taylor series expansion. The Taylor series expansion is applied to the activation function. The authors used a fifth-order expansion for Tanh ($f(x) = x + \frac{x^3}{3} + \frac{2x^5}{15}$), when $ \frac{x^3}{3} + \frac{2x^5}{15} <= 0.02$, $f(x) \approx x$. Solving the inequality results in $x>=0.39$ and Tanh$(2.90) \approx 1$. Therefore, only the values input $x$ in the range $0.39 - 2.90$ need to be stored in an LUT. Furthermore, authors in \cite{larkin2006efficient} makes use of minimax polynomial approximation. A minimax polynomial exists for every approximation method and can minimise the maximum error by evenly distributing the errors across the entire approximation range as opposed to the use of Taylor series expansion. The authors proposed the use of lower order polynomial by sub-dividing the input domain into smaller intervals using the Remez exchange algorithm. The Remez exchange algorithm is used to find approximate coefficients to generate minimax polynomials on discrete intervals \cite{larkin2006efficient}. This method requires the use of LUT, multiplier and adder.

In most cases, piecewise approximation requires one or more multipliers \cite{larkin2006efficient}. As described in \cite{courbariaux2014training,hubara2016binarized} multipliers are resource hungry and power-hungry devices on hardware. Other types of piecewise approximation have been shown not to require any multiplier but rather only comparators, multiplexers, shift operators, storage of several coefficients. Hard Tanh and sigmoid fall into these categories too. For a small degradation in performance accuracy for some problem space, hard and shift operators based piecewise approximation activation functions will suffice. 

FPGA implementation of LSTM has been explored in literature \cite{cao2019efficient,ferreira2016fpga, han2017ese,rybalkin2018finn}. LSTM's computational intensity has led to various ways of efficient implementation of LSTM. Weight pruning and compression are techniques that lower memory requirements and reduce complexity \cite{han2017ese}. Other complexity reduction methods are data representation and multiplier reduction through sparse LSTM. In literature, data representation in LSTM model go from 32 bits to binary \cite{rybalkin2018finn}. Commonly the activations and weights are represented with 8-bits or 16-bits \cite{cao2019efficient, ferreira2016fpga,han2017ese}. The use of the approximate multiplier \cite{azari2019energy,sim2017new} is on the increase in hardware neural networks. The approximate multiplier developed in \cite{sim2017new} and extended in \cite{azari2019energy}  includes hierarchical controllers that synchronize the variable cycle multiply operations with other single-cycle units. The principle is based on stochastic computing which allows for low cost and power but with increased fluctuation error and long latency. Its main component is a Finite State Machine (FSM) with $2^n$ states that generates a specific bit-stream.  The proposed design requires a variable number of clock cycles, which depend on the magnitude of the operands.  The advantages of the designs in \cite{azari2019energy,sim2017new}  are its low logic complexity, reduced power consumption and high error tolerance but with an increase in resource utilisation.Two popular works in literature have developed LSTM FPGA inference Engine namely C-LSTM \cite{wang2018c} and Efficient Speech Recognition Engine  (ESE) \cite{han2017ese}. The authors \cite{han2017ese}, build an ESE engine with sparse LSTM on FPGA. In the model, 16 multipliers were instantiated for element-wise multiplications per channel (total channel is 32).


\section{Implementation}
The algorithm (Equation~\ref{eq:concept}) can be adapted to generate both symmetric and asymmetric mappings\footnote{The manuscript that details the implementation is under a second review. This document will be updated as soon as the final link is known}.  

\subsection{Symmetric Activation Function - Square Law Nonlinearity (SQNL)}
Given a word size $R$, the symmetric activation function generator will use the following parameters:

\begin{itemize}
\item $C = 2^{R-2}$ and $M = 2^{R-1}$
\item $U(k) = \{-U_{MAX}, \cdots, U_{MAX}\}$ with $U_{MAX} = C = 2^{R-2}$. The length of $U(k)$ is a design decision.
\end{itemize}

With these parameters, the mapping can, analytically, be shown to be - 
\begin{equation}\label{eq:symmetric}
f(n) = \left\{
  \begin{array}{ll}
    -\frac{M}{2}						& : n < -M \\
    n + \frac{n^2}{2M} 	& : -M \le  n < 0 \\
    n - \frac{n^2}{2M} 	& : 0 \le n \le M \\
    \frac{M}{2} 					& : n > M
  \end{array}
\right.
\end{equation} 

\begin{figure}[h]
\centering
\includegraphics[]{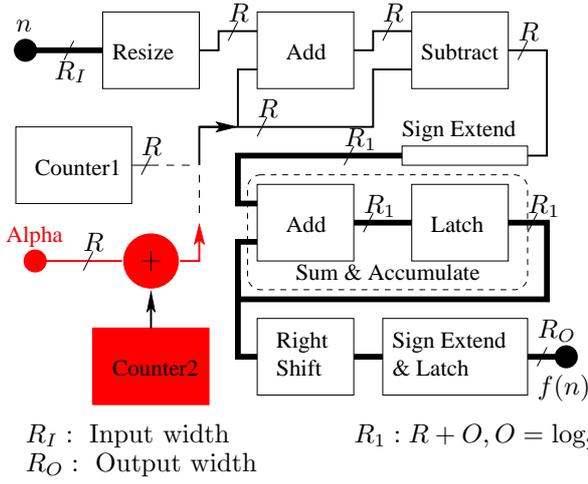}
\caption{\label{fig:full_schematic_sym_asym}Schematics of Symmetric and Asymmetric Activation Function Generator}
\end{figure}

Figure~\ref{fig:full_schematic_sym_asym} shows the implementation schematic and can be easily translated into HDL. The symmetric mapping requires Counter1 (alpha and Counter2 are required for the asymmetric generator).  The resolution of the GEMM operation will typically be of a higher resolution and hence may need to be resized before obtaining the non-linear mapping. 

Figure~\ref{fig:sqnl_sim_asymmetrical} shows the SQNL mapping with $R=8$ and $N = 8$. The SQNL is morphologically similar to the TanSig function. Although it is not symmetrical, the LogSQNL is also shown. It can be derived, in hardware, from the SQNL with almost zero hardware overhead because the divide by 2 (Div2) and offset reduce to 'shift' operations. 
  
\begin{figure}[h]
\centering
\includegraphics[width = 0.9\columnwidth]{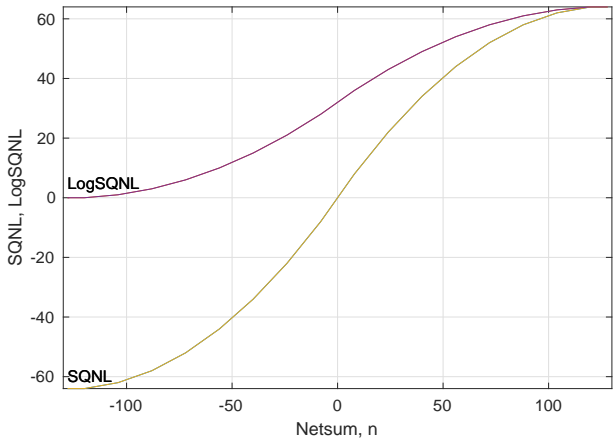}
\caption{\label{fig:sqnl_sim_symmetrical}Symmetrical mapping with $R=8$ and $N = 8$}
\end{figure}

With $R=8$ i.e. $-128 \le n < 127$, the output range is $-64 \le f(n) \le 63$ and $2 \le N \le 2^{R-1}$. The interative nature of Equation~\ref{eq:concept} implies a smaller value length of $U(k)$ is attractive. With $N = 2^{R-1}$ the mapping is perfectly smooth and perfectly . With $N < 2^{R-1}$ the mapping becomes piecewise linear. In Figure~\ref{fig:sqnl_sim_symmetrical} with $N = 8$ the mapping shows 16 linear segments that are evenly distributed across $-64 \le f(n) \le 63$. The deviation of the piecewise linear mappings from Equation~\ref{eq:concept} are developed.

\begin{figure}[h]
\centering
\includegraphics[width = 0.9\columnwidth]{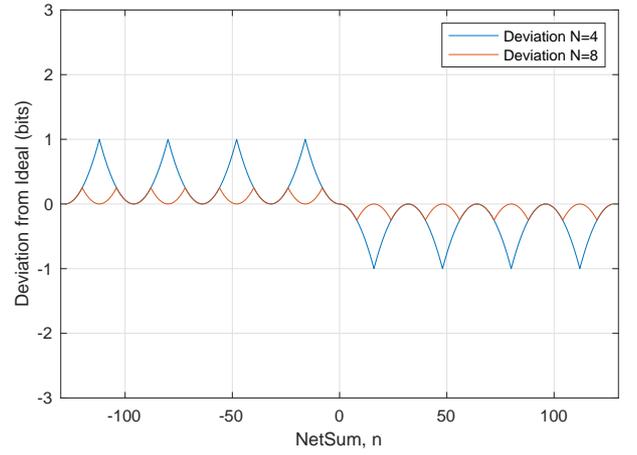}
\caption{\label{fig:deviationprofiles}Deviation from ideal for $R=8$ at $N = $ 4 and 8}
\end{figure}

Figure~\ref{fig:deviationprofiles} plots the deviations with $N = 4$ and $N = 8$. The deviation is at maximum when $n$ is equal to any of the values of $U(k)$ and zero midway between any two consecutive $U(k)$. As expected, the profile shows a reduction in deviation with an increase in $N$.  

\begin{figure}[]
\centering
\includegraphics[width=0.9\columnwidth]{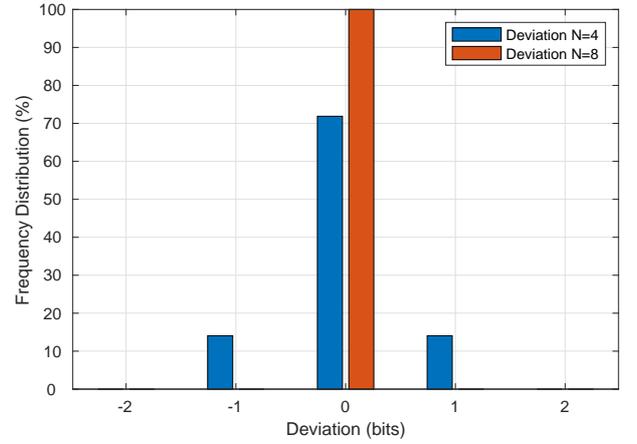}
\caption{\label{fig:deviation_histogram}Histograms of the deviation from ideal for $R=8$ at $N = $ 4 and 8}
\end{figure}

Figure~\ref{fig:deviation_histogram} shows the distribution of the deviations with $N = 4$ and $N = 8$. It shows that the deviations, for both, are at most $\pm 1$ bit away from the ideal mapping. With $N = 8$ the histogram shows that the deviations are within $\pm 0.25$. Theoretically, there will be a zero-bit error if $N = 8$ when $R = 8$. However, if $N$ is reduced to 4, more than 70\% of the range will still exhibit zero-bit errors.  

\subsection{\label{sec:sqsoft}Asymmetrical Activation Function}
Given a word size $R$, the asymmetric activation function generator will use the following parameters:
\begin{itemize}
\item $U(k) = \{-2U_{MAX}, \cdots, 0\} + \alpha$, where $0 \le \alpha \le \frac{M}{2}$
\item Adder saturates at $C = \{-U_{MAX}, M\}$,  where\\ $U_{MAX} = 2^{R-2}$ and $M = 2^{R-1}$,  i.e., only the lower saturation boundary is required
\end{itemize}

With these parameters, the mapping can, analytically, be shown to be -
\begin{equation}\label{eq:asymmetric}
f(n) = \left\{
  \begin{array}{ll}
      -\alpha					& : n < -\frac{M}{2} - \alpha \\
    \frac{(\frac{M}{2} + n + \alpha)^2}{2M} - \alpha	& : -\frac{M}{2} - \alpha \le n  \le \frac{M}{2} - \alpha\\
       n 					& : n > \frac{M}{2} - \alpha\\

  \end{array}
\right.
\end{equation} 

\begin{figure}[]
\centering
\includegraphics[width=0.9\columnwidth]{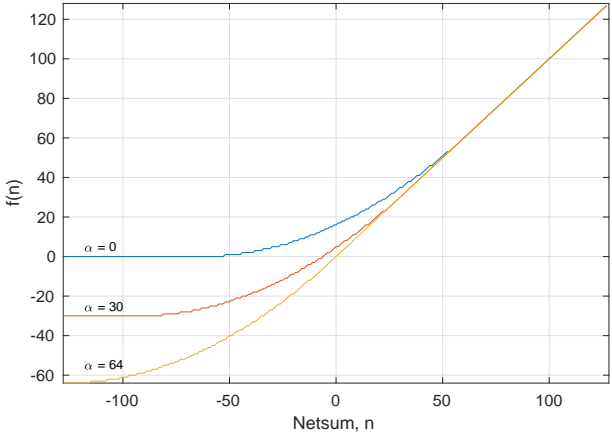}
\caption{\label{fig:sqnl_sim_asymmetrical}Asymmetrical mappings with $R=8$ and $N = 8$}
\end{figure}

A hardware implementation would require the 'alpha' and 'Counter2' elements in Figure~\ref{fig:full_schematic_sym_asym}. The asymmetrical mappings are shown in Figure~\ref{fig:sqnl_sim_asymmetrical}. With $\alpha= 0$, the mapping is similar to the SoftPlus function while with $\alpha > 0$ is similar to the ELU function. It should be highlighted that the mappings are not exponential and inherently quadratic.    

\subsection{\label{sec:gated}The Gated Activation (Scaled SQNL)}
\begin{figure}
\begin{center}
\includegraphics[]{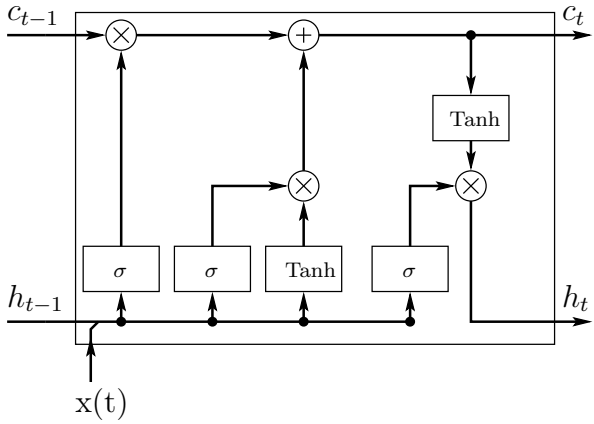}
\end{center}
\caption{A Conventional LSTM Cell}
\label{fig:lstmcell}
\end{figure}

LTSM cells, Figure~\ref{fig:lstmcell} require a bitwise scaling of a TangSig activation function with the output of an LogSig activation function. Each cell requires 3 multiplications. 

\begin{figure}[]
\centering
\includegraphics[]{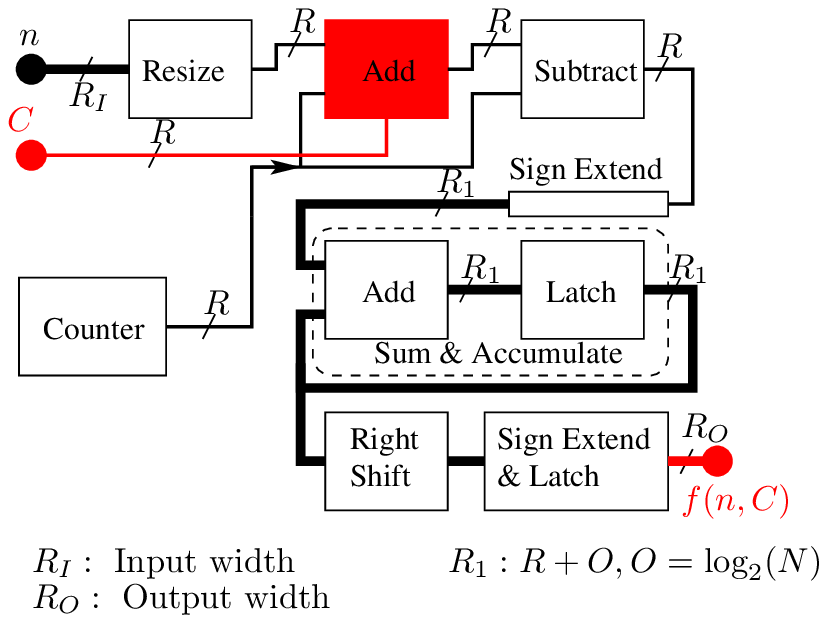}
\caption{\label{fig:full_schematic_gated}Schematic of the digital implementation of gated activation method.}  
\end{figure}

In the earlier sections we have shown an ability to make, in hardware, activation functions that are morphologically similar to both the TanSig and the LogSig. A modification to the SQNL schematic can integrate the scaling operation with a relatively small overhead in hardware resources. Specifically, the adder block is modified such that its saturation level is set by the required scaling. This is shown in Figure~\ref{fig:full_schematic_gated}. This, in-effect, achieves a scaling without requiring a physical multiplier. The mapping, Equation~\ref{eq:gated_analytic} has been derived.  

\begin{equation}\label{eq:gated_analytic}
f(n, C) = \left\{
  \begin{array}{ll}
    -C					& : n<-(U_{MAX} + C)\\
    n +  \frac{(-\bigtriangleup + n)^2}{4U_{MAX}} 	& : -(U_{MAX} + C) \le n < - \bigtriangleup\\
    n \frac{C}{U_{MAX}} 	& : -\bigtriangleup \le  n \le \bigtriangleup \\
    n -  \frac{(\bigtriangleup + n)^2}{4U_{MAX}} 	& : \bigtriangleup < n \le (U_{MAX} + C) \\
    C					& :n > (U_{MAX} + C)
  \end{array}
\right.
\end{equation}

\begin{figure}[]
\centering 
\includegraphics[width = 0.8\columnwidth]{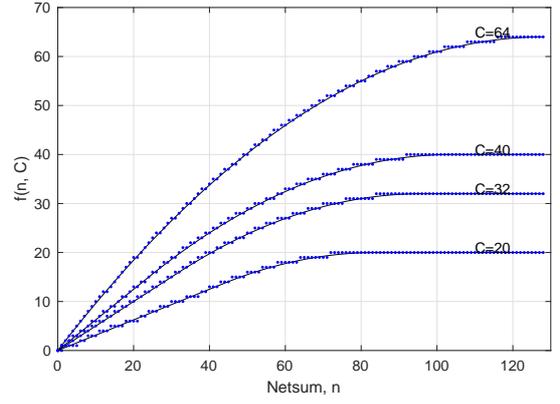}
\caption{\label{fig:gated_mappings}Gated Activations for $R = 8$, $N = 4$}
\end{figure}

At the limit, $C = 2^{R-2} = U_{MAX}$ and if $G(n) = f(n, 2^{R-2})$ then Equation~\ref{eq:gated_analytic} gives ${f(n,C) \approx m \times G(n)}$ where $m = C/U_{MAX}$. The range of the LogSQNL is $[0,2^{R-2}]$ and hence it can be connected to the $C$ port. This implements a scaling as required in the bitwise multiplication in LSTMs. The mappings of $f(n,C)$, with $R=8$ and $N = 8$, for different values of $n$ and $C$ are shown in Figure~\ref{fig:gated_mappings}. This figure also overlays Equation~\ref{eq:gated_analytic} with a solid black trace.

The algorithm as modelled by Equation~\ref{eq:gated_analytic} exhibits a small error from the ideal ${m \times G(n)}$. The plot with $C=64$ corresponds to ${G(n) = f(n, 64)}$. With $C = 40$, ${f(n,40) \approx \frac{40}{64} G(n)}$. Specifically at $n=40$, ${G(40) = 33.75}$, ${f(40,40) = 24}$ but ${\frac{40}{64} \times 33.75 = 21.09}$ and hence, the gated activation exhibits an error of $-2.91$. The error can be determined by Equation~\ref{eq:error_gated}.

\begin{figure}[]
\centering 
\includegraphics[width = 0.8\columnwidth]{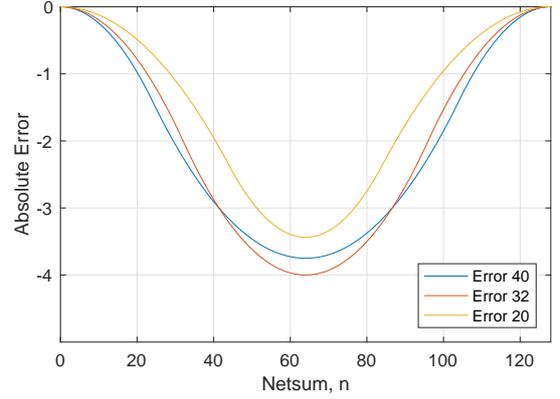}
\caption{\label{fig:error_gated}Error between $m \times G(n)$ and $f(n,C)$}
\end{figure}

\begin{equation}\label{eq:error_gated}
E(n,C) = \frac{C}{U_{MAX}}\times  G(n) - \left (n +  \frac{-(\bigtriangleup + n)^2}{4U_{MAX}}) \right )
\end{equation}

\begin{figure}
\begin{center}
\includegraphics[]{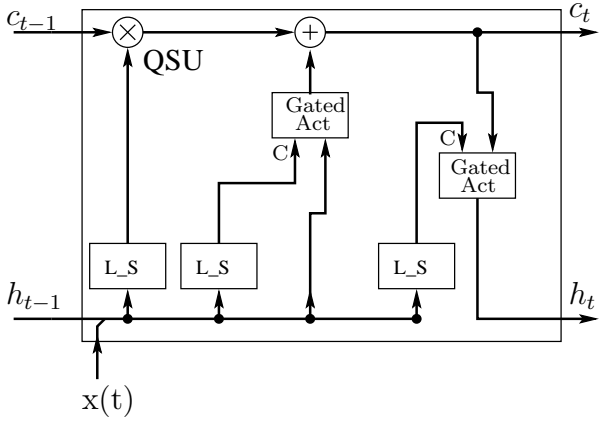}
\end{center}
\caption{An LSTM Cell using SQNL}
\label{fig:lstmcell_sqnl}
\end{figure}

Thus, an LSTM Cell can be implemented using the SQNL by replacing the TanSig and LogSig activation functions with the SQNL eqivalents. Additionally, the gated activation effectively eliminates two of the three multipliers as well. An LSTM Cell using the SQNL is shown in Figure~\ref{fig:lstmcell_sqnl}. In work,  we also propose a lower cost multiplier (QSU) that capitalizes on the reduced bit-widths. This eliminates all multipliers in the cell. 

\subsection*{\textbf{Remarks}}
\begin{enumerate}

\item The input range of the SQNL activation function (Equation~\ref{eq:symmetric}) is between $[-\infty,+ \infty ]$. The output is continuous and bounded. The SQNL family of functions are easily differentiable and satisfies the requirements of the universal approximation theorem \cite{hornik1989multilayer, leshno1993multilayer}. 

\item The derivatives are well-behaved and computationally attractive.

\item From a computational perspective, the quadratic is the simplist non-linearity. If a non-linear activation is required, the SQNL function is the most computationally efficient non-linearity for both inferencing and learning on ASICs, FPGAs, or CPUs.

\item The benefits of an hardware implementation of SQNL can be realised with or without the availability of hardware multipliers.  

\end{enumerate}

\section{Resource Utilization}

The SQNL algorithm can be implemented using standard building blocks. If hardware multipliers are available,  the closed-form quadratic equations can be implemented instead. 

The real-estate on silicon is dependent on the target chip technology. The metrics shown here has based on a an Intel Cyclone V compiled using Quartus Prime ver 18.1. On FPGAs, the basic building block is vendor dependent and hence we have also attempted to offer a vendor independent metric. To do this, we assume that the basic building block is a NAND gate. With this assumption, we estimate the gate usage for the relevant building blocks and shown in Table~\ref{tab:genericGateUsage}. The SQNL blocks can then be quantified using these building blocks. These indicative metrics will be used in all subsequent comparisons. 
 
\begin{table}[h]
\centering
\caption{\label{tab:genericGateUsage}Indicative Gate Usage}
\begin{tabular}{lccc}
\hline 
Digital Block     & Cell/bit & Gates/bit     & Total Gate \\ \hline
Single register    & - & -  & 4          \\ 
1bit full adder   & - & -     & 9          \\
8bit adder      & - & -      & 72         \\ 
9bit adder      & - & -      & 81         \\
8bit 2's Complement & - & -   & 80         \\
9bit 2's Complement & - & -   & 90         \\ 
2 to 1 mux (1bit) &- &3 &3 \\ 
LUT 8 bits (1-sided) & 127 & 381   & 2667         \\
LUT 12 bits (1-sided)& 2047 & 6141   & 67551         \\ 
LUT 8 bits (2-sided)& 255 & 765   & 6120         \\
LUT 12 bits (2-sided)& 4095 & 12285   & 147420         \\ 
Booth 8 bits &-  &   - & 754         \\
Booth 12 bits & - & -   & 1124         \\ \hline 
\end{tabular}
\end{table}

The comparison of the SQNL was undertaken with competing techniques: LUT, PWL and multipliers. To determine the gate usage of a multiplier, we use a custom implementation of a radix 4 multiplier and coded in HDL. The  gate usage is also listed in Table~\ref{tab:genericGateUsage}.

\begin{table*}[h]
\centering
\caption{\label{tab:logsqnl}Resource utilization of symmetric and asymmetric functions implementation using a custom Booths Radix-4 multiplier (mult), multi-clock (counter, $N = 8$), and LUT solution. As displayed the counter based method for the two function types consistently outperform the multiplier solution on both ASIC and FPGA platforms. At a lower resolution, the LUT performs slightly better than the counter solution on FPGA but worse on ASIC when compared with the counter solution. The LUT on the other hand is only better for lower resolution and can not accommodate applications where higher resolution is required. The counter solution scales well across different resolutions.}
\begin{tabular}{lllllllllll}
\hline
                    &         & \multicolumn{3}{l}{Symmetric} & \multicolumn{6}{l}{Asymmetric}                                                        \\
                    &         & \multicolumn{3}{l}{SQNL}      & \multicolumn{3}{l}{SQLU ($\alpha=2^{R-2}$)} & \multicolumn{3}{l}{SQ\_Softplus ($\alpha=0$)} \\
R                   & Methods & Gates      & ALM     & FF     & Gates         & ALM        & FF        & Gates           & ALM          & FF          \\ \hline
\multirow{3}{*}{8}  & Mult    & 963           &  120       & 63       &  987             &           124 &   63        &  826               &   100           &  54           \\
                    & LUT     & 2779           & 17        &  8      &    2779           &           17 &  8         &  2747               &    19          &   8          \\
                    & Counter &388            &  34       &  24      & 460              &           31 & 22           &  460               &   29           &   24          \\ \hline
\multirow{3}{*}{12} & Mult    & 1400           &  180       &  93      &   1473            &           187 &    93       &  1232               &    153          & 80            \\
                    & LUT     & 67719           & 234        &  12      &      67719         &           234 & 12          &   67671              &      125        &    12         \\
                    & Counter & 556           &  46       &  31      &    664           &           41 & 31          &  664               &   39           &    30        
 \\ \hline
\end{tabular}
\end{table*}

A detailed analysis of the implementation is discussed in a manuscript that is under second review. Below we summarize our comparisons of the SQNL with competing techniques. 
\begin{itemize}
	\item Counter - At low/mid resolutions the counter based solution gives the maximum density of activation functions. The counter-based solution permits the encapsulation of symmetric and asymmetric functions into a single entity. Thus activation functions can be dynamically adapted during training and inferencing. Our experiments on various models show that the absolute error due to $N\le8$ has no impact on model performance. However if $R > 16 \mbox{ and } N > 16$, the multiplier may be more efficient but if $N \le 8$ the counter based method is much more attractive.   
	
	\item Multiplier - A multiplier solution would be attractive if free multipliers are available and if high/full precision is essential. Furthermore, since our square-law based functions are not an approximation, the direct solution will not introduce any approximation errors when compared to the PWL methods found in the literature. The conditionals in Equation~\ref{eq:symmetric} and \ref{eq:asymmetric} can be implemented using combinatorial logic. This will lead to addition timing overhead from propagation delay but will not require additional clocks.  
	
	\item LUT - A LUT based function underperforms compared with a counter-based solution. At higher resolutions, the LUT is not a practical option due to excessive gate usage. At lower resolutions ($R<8$), a LUT may outperform a counter-based solution on an FPGA but this is device-dependent. 

\end{itemize}

The resource usage implications for an LSTM cell have also been detailed. In summary, 
\begin{itemize}
	\item Piecewise Linear Approximation -
	Each of the 5 activation functions will require 5 multipliers and an additional 3 multipliers for bitwise operations. Assuming 3 multiplications per DSP would imply a minimum of 3 DSPs but the fitter reports usage of 7 DSPs. Increasing the resolution to 16 bits should result in 4 DSP blocks but again the fitter reports only 7 DSPs. It is possible that the data flow necessitates the allocation of multipliers in individual DSPs.
	
	\item Lookup Table -  
	With this method, the table values are stored in the LUT's within the ALMs. The three bitwise multipliers would ideally require a single DSP but the fitter reports 2 DSPs. As expected, the 16-bit cell shows a big increase in ALM usage to store the table also requires 2 DSPs. As with the PWL method, the data flow necessitates the allocation of multipliers in individual DSPs.
	
	\item Hard Tanh and Hard Sigmoid -
	The hard activations are the simplest to implement and the bulk of the resource usage will be used to effect the saturation logic at the breakpoints. As with the Lookup Table method, bitwise multipliers require 2 DSPs for both 8 and 16-bit resolutions.    
	
\end{itemize}

\begin{table*}[h]
\centering
\caption{\label{tab:resources2}Resource utilization of  LSTM cell}
\begin{tabular}{lccccccc}
\hline
\multirow{2}{*}{Method}                                          & \multicolumn{3}{l}{8 bits} & \multicolumn{3}{l}{16 bits} \\
                                                                 & ALM   & FF   & DSP  & ALM   & FF   & DSP    \\ \hline
Piecewise Activation \cite{wang2018c}        & 81   & -         & 7    & 107   & -        & 7     \\
LUT \cite{han2017ese}  & 96    & 16          & 2    & 6,568   & 32         & 2     \\
Hard Activation  & 20    & -          & 2    & 53   & -         & 2    \\
Gated Activation (Proposed)                                                    & 52    & 55          & -    & 150   & 175         & -   \\

\hline
\end{tabular} 
\end{table*}

\section{SQNL Family for Simulations}
The mappings developed for hardware implementation may have a restrictive use in simulation frameworks. These quadratic mappings can be translated to suit CPUs implementations. 

\subsection{SQNL}
Equivalent to the TanSig (TanH) mapping - 
\[
	f(n) = \left\{
  	\begin{array}{ll}
    1           & : n > 2.0\\
    n - \frac{n^2}{4}   & : 0 \le n \le 2.0\\
    n + \frac{n^2}{4}   & : -2.0 \le  n < 0\\
    -1          & : n < -2.0
  	\end{array}
	\right.
\]

\subsection{SQ\_LogSig}
Equivalent to the Sigmoid mapping - 
\[
	f(n) = \frac{f_{SQNL}}{2} + 0.5  
\]

\subsection{SQLU}
Equivalent to the ELU mapping - 
\[
f(n) = \left\{
  \begin{array}{ll}
    n 	& : n >0\\
    n + \frac{n^2}{4} 	& : -2.0 \le  n \le 0\\
    -1					& : n < -2.0
  \end{array}
\right. 
\]

\subsection{SQ\_SOFTMAX}
Equivalent to the Softmax mapping - 
\[
f(n) = \left\{
  \begin{array}{ll}
    n 	& : n >0.5\\
    \frac{(n+0.5)^2}{2} 	& : -0.5 \le  n \le 0.5\\
    0					& : n < -0.5
  \end{array}
\right.
\]

\subsection{SQ\_SQISH}
Equivalent to the SQISH mapping -
\[
f(n) = \left\{
  \begin{array}{ll}
    n+\frac{n^2}{32} 	& :n>0\\
    n + \frac{n^2}{2} 	& : -2.0 \le  n < 0\\
    0					& : n < -2.0
  \end{array}
\right.
\]

\subsection{SQ\_REU}
Equivalent to the REU mapping -
\[
f(n) = \left\{
  \begin{array}{ll}
    n 	& : n >0\\
    n + \frac{n^2}{2} 	& : -2.0 \le  n \le 0\\
    0					& : n < -2.0
  \end{array}
\right.
\]

\section{Benchmark Tests}
The mappings have been written for ANN engines on a PC (Python and Matlab) and their performances compared. In \cite{wuraola2018sqnl, wuraola2021efficient}. the authors have compared a variety of datasets ranging from tiny to large. By-and-large, the morphology has minimal impact on performance. Conservatively speaking, the performance of the quadratic morphology is comparable with the exponential equivalents. However, with reservation we can also state that the performance of the quadratic functions is better - at least in some tests. The authors acknowledge that fine tuning of every hyperparameter will, without doubt, result in differing judgements.  

\section{Conclusions}
The SQNL algorithm has been introduced. The extensions of the basic algorithm to construct asymmetrical and scaled activations have also been introduced. Their underlying equations have been presented and errors arising from the implementation decisions have been shown. This document suggests that the SQNL algorithm has demonstrable benefits for ANN engines.

This document also redirects interested viewers to the authors detailed manuscripts that have been peer reviewed.



\end{document}